\documentclass[letterpaper, 10 pt, conference]{ieeeconf}  % Comment this line out if you need a4paper

\IEEEoverridecommandlockouts                              % This command is only needed if 
\usepackage{amsmath}
\usepackage{graphics} % for pdf, bitmapped graphics files
\usepackage{graphicx}
\usepackage{epsfig} % for postscript graphics files
\usepackage{cite}
\usepackage{xcolor}
\usepackage{comment}
\usepackage{soul}
\usepackage{adjustbox}
\usepackage{multirow}
\usepackage{wasysym}
\usepackage{colortbl}
\usepackage{algorithm} % Algorithm with lines
\usepackage{algpseudocode}
\usepackage{amssymb}
\usepackage{nicefrac}
\usepackage[switch]{lineno}
\usepackage{soul}
\usepackage{colortbl}

\usepackage{nicefrac}       % compact symbols for 1/2, etc.
\usepackage[nopatch=eqnum]{microtype}
\usepackage{caption}
\usepackage[font={scriptsize}]{caption}
\usepackage{hyperref}
\hypersetup{
    colorlinks = true,
    citecolor = {green},
    linkcolor  = {blue},
}

\usepackage{multirow}
\newenvironment{myitem}{\begin{list}{$\bullet$}
{\setlength{\itemsep}{-0pt}
\setlength{\topsep}{0pt}
\setlength{\labelwidth}{0pt}
\setlength{\leftmargin}{10pt}
\setlength{\parsep}{-0pt}
\setlength{\itemsep}{0pt}
\setlength{\partopsep}{0pt}}}%
{\end{list}}

\renewcommand{\baselinestretch}{0.993}

\definecolor{todo}{rgb}{1.0, 0., 0.}
\definecolor{flodarkpurple}{rgb}{0.288,0.1196,0.7}
\definecolor{bin}{rgb}{0.9,0.,0.5}
\definecolor{han}{rgb}{0.08,0.33,0.6}
\definecolor{stefan}{rgb}{0.,0.,0.9}

\renewcommand\hl[1]{#1}  % Diable highlight
\newcommand{\coolname}{\textit{FuncGrasp}}
\newcommand{\authorhref}[3][flodarkpurple]{\href{#2}{\color{#1}{#3}}}

\usepackage[capitalize]{cleveref}
\crefname{section}{Sec.}{Secs.}
\Crefname{section}{Section}{Sections}
\Crefname{table}{Table}{Tables}
\crefname{table}{Tab.}{Tabs.}

\title{\LARGE \bf \coolname: Learning Object-Centric Neural Grasp Functions \\ from Single Annotated Example Object}
\author{
\authorhref{https://hanzhic.github.io/}{Hanzhi Chen}$^{1,3}$ \quad 
\authorhref{https://binbin-xu.github.io/}{Binbin Xu}$^{2}$ \quad 
\authorhref{https://srl.cit.tum.de/}{Stefan Leutenegger}$^{1,3}$ 
\thanks{This work was funded by TUM Georg Nemetschek Institute under the project SPAICR. }   
\thanks{\scriptsize $^{1}$ Smart Robotics Lab, School of CIT, Technical University of Munich. 
{\tt\small \{hanzhi.chen, stefan.leutenegger\}@tum.de}. }
\thanks{\scriptsize $^{2}$ University of Toronto Robotics Institute, University of Toronto. 
{\tt\small binbin.xu@utoronto.ca}}
\thanks{\scriptsize $^{3}$ Munich Institute of Robotics and Machine Intelligence (MIRMI).}
}
\begin{document}

\maketitle
\thispagestyle{empty}
\pagestyle{empty}

%%%%%%%%% ABSTRACT
\begin{abstract}

We present \coolname{}, a framework that can infer dense yet reliable grasp configurations for unseen objects using one annotated object and single-view RGB-D observation via categorical priors. Unlike previous works that only transfer a set of grasp poses, \coolname{} aims to transfer \textit{infinite} configurations parameterized by an object-centric continuous grasp function across varying instances. To ease the transfer process, we propose \textit{Neural Surface Grasping Fields} (NSGF), an effective neural representation defined on the surface to densely encode grasp configurations. Further, we exploit function-to-function transfer using sphere primitives to establish semantically meaningful categorical correspondences, which are learned in an unsupervised fashion without any expert knowledge. We showcase the effectiveness through extensive experiments in both simulators and the real world. Remarkably, our framework significantly outperforms several strong baseline methods in terms of density and reliability for generated grasps.

\end{abstract}
\section{Introduction}
\label{sec:intro}

When a robot is interacting with the physical world, inferring suitable grasping strategies for objects of interest has been a long-standing problem in the robotics community. In recent years, learning-based methods have significantly boosted the performance of stable grasp detection relying on supervision from a massive amount of training data \cite{johns2016deep, liang2019pointnetgpd, mousavian20196, ten2017grasp, wen2022catgrasp, weng2023neural}. Nevertheless, generating dense yet reliable grasp configurations from these methods usually presents a challenge since rigorous filtering criteria are employed to select the most confident grasp proposals from (partial) observations. This tends to ignore the variability in workspace setups, leading to practically unreachable grasps due to kinematic constraints. Several attempts have been made to address this seemingly “mutually exclusive” process. One line of research explores using categorical priors to generate rich grasp configurations for novel objects. At its core, it first (densely) annotates one source object, then establishes categorical correspondences among instances, which can be realized by learning feature-metric descriptors\cite{simeonov2022neural, florence2018dense}, deformation fields \cite{wen2022transgrasp}, etc. They have demonstrated a substantial potential for few-shot grasp learning. However, as the unseen target objects' inferred grasp density is completely dependent on the size of the set of discrete grasp poses to transfer, it could be impractical to transfer a huge amount of configurations since they demand gradient-based optimization per grasp (e.g., high-dimensional descriptor matching loss in \cite{simeonov2022neural}). Another trend of research leverages neural networks to create smooth grasp representations not restricted by resolution. Though existing methods like \cite{weng2023neural} have demonstrated impressive results in terms of density and reliability, massive datasets are required for better generalization. Moreover, function-level transfer for grasps is often achieved through shallow vector embeddings \cite{weng2023neural, blukis2022neural}, which can potentially compromise expressiveness. This limitation is evident in several 3D shape modeling works with even more constrained dimensionality \cite{xu2022learning, wang2021dsp}.

\begin{figure}[t!]
    \centering
    \includegraphics[width=0.48\textwidth ]{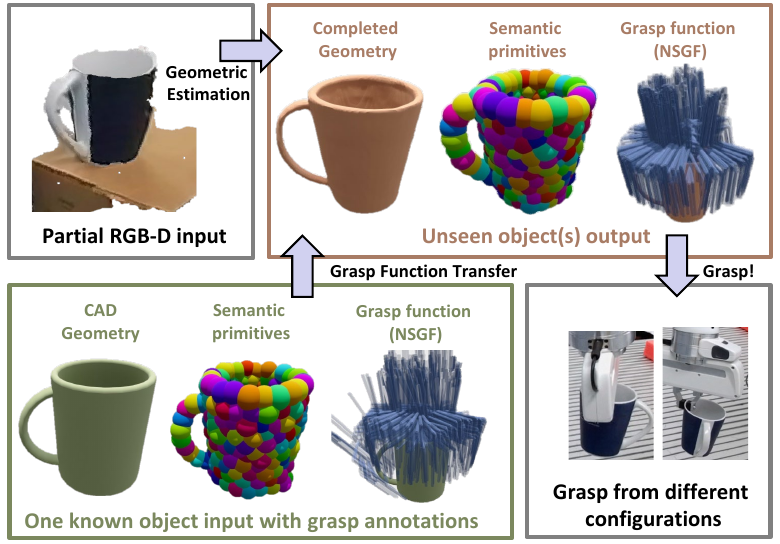}
    % \vspace{-1.7\baselineskip}
    \caption{Given a partial RGB-D input, our framework transfers a known object's continuous grasp function fitted from discrete annotations to the unseen object. We represent such a function using our proposed \textit{Neural Surface Grasping Fields} (NSGF) formulation. This process is achieved by completing the object's geometry and estimating its semantic primitives learned in an unsupervised fashion. Using the transferred NSGF, the robot can query the dense dependable grasp knowledge embedded in a smooth function to conduct grasping from different configurations.}
    \label{Fig:teaser}
    \vspace{-2.0\baselineskip}
\end{figure}

Given all these emerging challenges, our goal is to develop a data-efficient framework that enables robots to infer a wide range of dependable grasp configurations for previously unseen objects by transferring dense grasp knowledge from one single annotated object within the same class. We seek to address two key questions: (1) how to effectively represent dense grasp configurations; and (2) how to accurately transfer such grasp representations, tailored for unseen objects. To this end, we parameterize grasp configurations as object-centric functions, leveraging the power of neural representations to smoothly interpolate between discrete samples. Inspired by \cite{sundermeyer2021contact}, we first propose \textit{Neural Surface Grasping Fields} (NSGF) to continuously define grasp configurations on the entire surface. The intuition for NSGF lies in that the grasp pose defined on $\mathrm{SE}(3)$ has much higher dimensions compared to implicit shape representations, and only a very small portion of the object's volume has meaningful results. Hence, we decouple shape and grasp modeling in contrast to previous methods, e.g., \cite{jiang2021synergies, blukis2022neural}, by first obtaining the completed geometry of the object from \cite{xu2022learning}, then extracting the object surface so that grasp configurations can be effectively embedded in a more bounded space. This also eases the function transfer process thanks to the explicit parameterization of geometry. Moreover, we leverage semantically consistent sphere primitives learned in an unsupervised fashion based on \cite{hao2020dualsdf} to achieve the transferability of NSGF, respecting the fact that grasp configurations shall not vary much in local surface regions.  An overview of our proposed framework is shown in Fig.~\ref{Fig:teaser}. To summarize, our key contributions are:
\begin{itemize}
    \item An effective neural representation to encode grasp configurations for surface points, \textit{Neural Surface Grasping Fields} (NSGF), capable of harnessing geometric cues to provide accurate, reliable, and dense grasp poses.
    \item A novel approach to perform function-level transfer for our proposed NSGF, leveraging semantic primitives learned in an unsupervised manner. To the best of our knowledge, we are the first to design a feasible paradigm to transfer continuous grasp functions instead of using shallow vector embeddings. 
    \item Extensive experiments in simulators and the real world to validate the effectiveness of our framework, \coolname{}.
\end{itemize}

\section{Related works}
\label{sec:related}
\begin{figure*}[t!]
    \centering
    \includegraphics[width=0.9\textwidth]{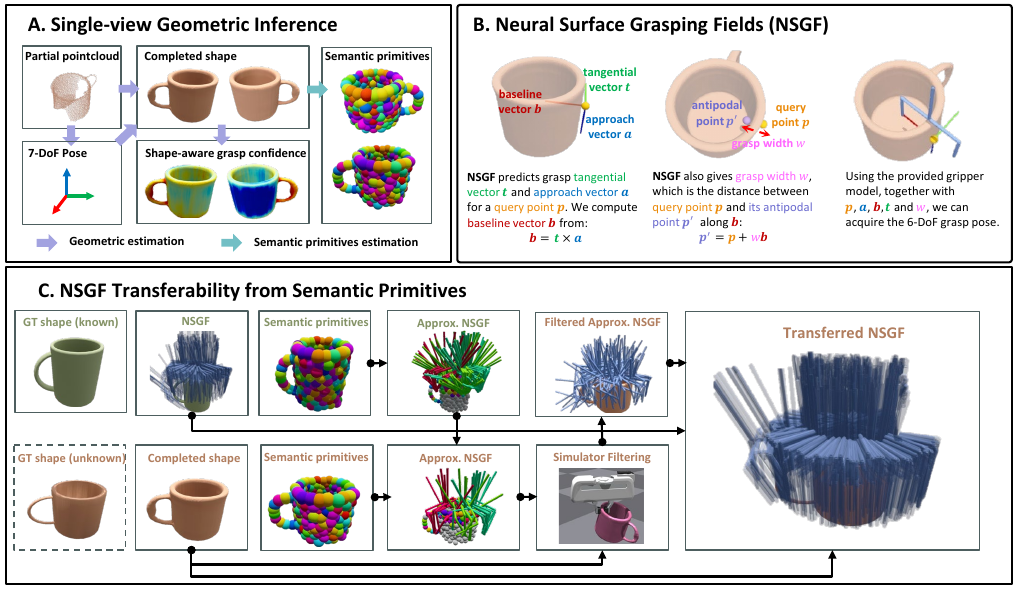}
    \vspace{-0.3\baselineskip}
    \caption{Illustration of our proposed framework, \coolname{}. (A) Our geometric estimation module infers the target object's 7-DoF pose, completed shape, semantic primitives, and shape-aware confidence (red indicates low; blue indicates high). (B) Our Neural Surface Grasping Field (NSGF) formulation defines point-wise grasp configurations on the surface. (C) We approximate NSGF using semantic primitives. We sample a small number of grasp configurations for each primitive and transfer them to the target object using the corresponding primitive (grippers colored magenta, green, and cyan indicate three different primitives). After adjusting the transferred grasps based on the object's shape and filtering invalid samples in the simulator, we fit a new NSGF using the rest of the samples to achieve grasp function transfer.  }

    \label{Fig: pipeline}
    \vspace{-1.9\baselineskip}
\end{figure*}

\noindent \textbf{Model-free grasp detection.} Without the requirement for CAD models, model-free grasp detection methods aim to produce dense point(pixel)-wise predictions on grasping quality and poses from sensor observations. GPD \cite{ten2017grasp} and PointNetGPD \cite{liang2019pointnetgpd} leverage deep neural networks to predict the scores of grasp pose candidates obtained from the pointcloud. VGN \cite{breyer2020volumetric} is fed with a TSDF representation of the scene and in turn, outputs structured voxelgrids with grasp quality, orientation, and width. 6-DoF GraspNet \cite{mousavian20196} models grasp detection as a generative process and uses a variational auto-encoder (VAE) to produce grasp proposals. 
Contact-GraspNet \cite{sundermeyer2021contact} eases the grasp learning process by proposing a more structured grasp representation. 
Notably, those methods usually require huge amounts of data with grasp annotations for supervision, which can be time-consuming even when employing advanced physics simulators. Our method only requires annotations for one object from each category, eliminating the need for large-scale simulations to acquire grasp labels.

\noindent \textbf{Category-level grasp learning.} Another line of research conducts model-free grasp learning from an object-centric perspective. Those works usually assume consistent categorical priors among objects to establish meaningful correspondence. CaTGrasp \cite{wen2022catgrasp} proposes non-uniform normalized object coordinates for better correspondence learning so as to build grasp codebooks in canonical space. DON \cite{florence2018dense} and NDF \cite{simeonov2022neural} leverage neural networks to learn deep descriptors to discover categorical correspondence in a self-supervised manner. kPAM \cite{manuelli2019kpam} establishes categorical correspondence through learning manually annotated 3D key points. TransGrasp \cite{wen2022transgrasp} proposes to use implicit deformations fields to infer grasps for novel objects from a pre-labeled instance. Instead of transferring a pre-labeled set of discrete grasp poses as in previous works, our approach aims to infer a continuous grasp function for novel objects, which provides dense yet reliable configurations and is not limited by resolution.

\noindent \textbf{Neural representations in grasping.} Recently neural representations have shown a strong capacity for novel view synthesis \cite{mildenhall2020nerf,chen2023texpose,martinbrualla2020nerfw}, 3D reconstruction \cite{park2019deepsdf,mescheder2019occupancy,chibane2020ndf}, and generative modeling \cite{Niemeyer2020GIRAFFE,jo2021cg}. At their core, multi-layer perceptrons (MLP) are used to encode the scene content of every 3D point. In the context of robotic grasping, such scene contents can be represented using geometric information (density, occupancy, (un-)signed distance), deep descriptors, and grasp configurations. Most attempts \cite{ichnowski2021dex,kerr2022evo,yen2022nerf} use neural representations to learn the accurate geometry of photometrically challenging objects with depth sensing failures. \cite{simeonov2022neural,chun2023local} try to use neural fields to discover distinctive descriptors. NGDF \cite{weng2023neural} models the grasp distribution as a distance field. \cite{blukis2022neural} uses neural radiance fields (NeRF) to unify the learning process of shape, appearance, and grasp. As the grasp poses are distributed in unconstrained \text{SE}(3) space and tend to lie near the surface, in our NSGF formulation, we propose to define them explicitly on the surface instead of the volume so as to increase the function's expressiveness and ease the transfer process.

\noindent \textbf{Neural fields transfer.} With the prevalence of neural representations, a few works started to explore approaches to conduct function-level transfer. NeSF \cite{vora2021nesf} introduces an approach to transfer neural density fields to semantic fields through voxel grids-based field approximation. NFGP \cite{yanggeometry} conducts geometry processing tasks, e.g., shape deformation, directly on neural signed distance fields through invertible neural networks. Inspired by NeSF \cite{vora2021nesf}, we leverage sphere primitives possessing semantic consistency to approximate our proposed NSGF to achieve flexible field transfer.

\section{Methods}
\label{sec:method}

Assuming one source object with one grasp function pre-fitted to annotated grasp labels that can be obtained manually or from simulations, our objective is to conduct grasp function transfer to several unseen target objects of the same type respecting their geometric features. For each target object, our pipeline only requires a partial pointcloud extracted from a single-view RGB-D frame. We introduce geometric inference to predict completed geometry, shape-aware grasp confidence, and semantic primitives in Sec. \ref{Sec:geoinf}. In Sec. \ref{Sec:trans}, we explain the proposed approach that uses primitive-based shape abstraction to transfer our formulated object-centric grasp function presented in Sec. \ref{Sec:nsgf}. The implementation details are provided in Sec. \ref{Sec:implement}. Detailed illustration of our framework is provided in Fig.~\ref{Fig: pipeline}.

\subsection{Single-view Geometric Inference}\label{Sec:geoinf}
\noindent \textbf{Geometric estimation.} Given a single-view RGB-D image input together with a foreground mask for the object of interest provided by an off-the-shelf detector \cite{kirillov2020pointrend}, we acquire a segmented pointcloud $\mathcal{X}=\{\mathbf{x}^1,...,\mathbf{x}^{N_{\mathcal{X}}}\}$ and its voxelized TSDF volume $\mathbf{V}$ for each object in the scene. We further feed $\mathcal{X}$ to a pre-trained pose estimator \cite{li2023generative} to acquire its 7-DoF pose $\mathbf{T} \in \mathrm{SIM}(3)$ from its canonical system to the camera system. $\mathbf{V}$ is then canonicalized with $\mathbf{T}$ and passed to the shape completion module from \cite{xu2022learning} to acquire the complete geometry represented by a voxel grid filled with occupancy probabilities $\mathbf{O}$. The object mesh $\mathcal{M}$ is extracted from $\mathbf{O}$ using multi-resolution iso-surface extraction strategy from \cite{mescheder2019occupancy}. We can also compute the shape confidence for mesh surface point $\mathbf{p}$ represented by the norm of their gradients w.r.t the occupancy probabilities, i.e., $||\partial \mathbf{O}[\mathbf{p}]/\partial \mathbf{p}||_2$. Shape confidence will later be used to rank grasp poses decoded from our formulated object-centric grasp function.

\noindent \textbf{Unsupervised semantic primitives learning.} Inspired by recent advances in shape manipulation \cite{hao2020dualsdf}, we establish the categorical correspondence by exploring the shape abstraction in a semantically consistent fashion. Specifically, an object's geometry can be approximated by a fixed number of spherical primitives and this abstraction leads to part-to-part correspondence 
(c.f. color-coded semantic primitives in Fig. \ref{Fig: pipeline}-C). More importantly, such correspondence labels can be learned in an unsupervised manner without any expert knowledge in contrast to \cite{manuelli2019kpam}. After fitting the primitive-based representation for each object used for training, we assign the closest primitive label to every surface point. During inference, we use a part segmentation network to estimate point-wise semantic primitive labels for uniformly sub-sampled surface points. 
For points with the same label, we pass them to Mean Shift to compute the clustering center, which is the final predicted primitive center location. All primitive centers are denoted as $\mathcal{S} = \{\mathbf{s}^1,...\mathbf{s}^{N_{\mathcal{S}}}\}$, where $N_{\mathcal{S}}$ is the number of pre-defined primitives. Here we obtain a geometric representation $\mathcal{G}$ for each unseen object with a tuple  $\mathcal{G} = (\mathbf{T}, \mathcal{M}, \mathcal{S})$. Note $\mathcal{M}, \mathcal{S}$ are represented in the canonical system.  

\subsection{Neural Surface Grasping Fields (NSGF)}\label{Sec:nsgf}
\begin{figure}[t!]
    \centering
    \includegraphics[width=0.9\columnwidth]{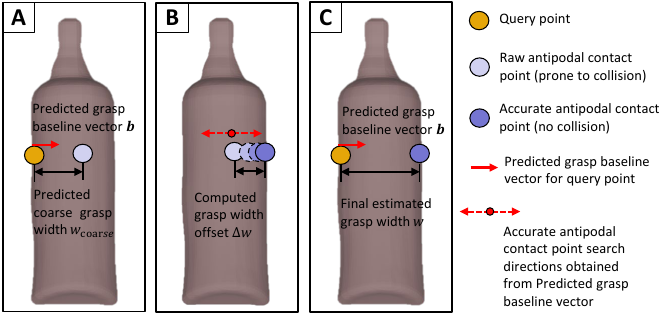}
    \caption{(A) Raw antipodal contact point using predicted coarse width $w_\text{coarse}$ without geometric awareness, leading to grasp failure due to collision. (B) Use of completed geometry to search for nearby on-surface points. (C) Accurate antipodal contact point thanks to precise shape completion, yielding a collision-free grasp pose.}
    \label{Fig: width}
    \vspace{-1.8\baselineskip}
\end{figure}

\noindent \textbf{NSGF formulation.} We propose to use a continuous function to represent grasp configurations. This function is parameterized by an MLP and defined on the object surface to map the point to its grasp validity and grasp pose. We refer to such representation as \textit{Neural Surface Grasping Fields} (NSGF). As shown in Fig.~\ref{Fig: pipeline}-B, inspired by\cite{sundermeyer2021contact}, for each point $\mathbf{p}$, NSGF outputs its grasp validity $q$, grasp approach vector $\mathbf{a}$, grasp tangential vector $\mathbf{t}$ and grasp width $w$. Grasp baseline vector $\mathbf{b}=\mathbf{t}\times\mathbf{a}$. For each point, the grasp pose can be solved using its coordinate $\mathbf{p}$, the predicted width $w$, rotation vectors $\mathbf{a}$, $\mathbf{b}$, $\mathbf{t}$, and the gripper model. We notice that such representation often fails when dealing with thick objects like bottles as the predicted width has no geometric awareness. Different from \cite{sundermeyer2021contact}, we harness the completed geometry to acquire a more precise grasp width from raw prediction $w_\text{coarse}$. As shown in Fig.\ref{Fig: width}-A, the antipodal point $\mathbf{p}^{\prime}$ is initialized along the baseline vector $\mathbf{b}$, i.e., $\mathbf{p}^{\prime}=\mathbf{p} + w_{\text{coarse}}\mathbf{b}$. Then we search for width offset $\Delta w$ so that $\mathbf{p}^{\prime}$ can be moved to the nearby surface along $\mathbf{b}$, i.e., $\mathbf{p}^{\prime}=\mathbf{p} + (w_{\text{coarse}} + \Delta w) \mathbf{b}$, and the final grasp width is $ w=w_{\text{coarse}} + \Delta w$ (\textit{c.f.} Fig.~\ref{Fig: width}-B,C). In practice, we query the occupancy value to determine if a point lies on the surface. 
With a slight abuse of notation, we denote NSGF as:
\begin{equation}
\begin{aligned}
    F(\mathbf{p}) \colon S^2 \to \mathbb{R} \times \mathrm{SE}(3) . 
\end{aligned}
\end{equation}
Notably, NSGF is defined directly on the 2D surface space instead of bounding volume as \cite{jiang2021synergies, blukis2022neural} and thus avoids extensive forward-passing and post-processing to locate stable grasps inherently near the surface.

\noindent \textbf{NSGF fitting.} To fit an NSGF $F$ for one object, validity loss ($l_{\text{v}}$) and rotation vectors loss ($l_{\text{r}}$) are adopted from \cite{sundermeyer2021contact}. Note in $l_{\text{r}}$, we regress the tangential vector instead of the baseline vector as we empirically find it helps encode more stable grasp poses. With the ground-truth antipodal contact point $\mathbf{p}^{\prime}_{\text{gt}}$, coarse width regression loss for each valid point is:

\begin{equation}
\begin{aligned}
l_{\text{w}}=\sum_{w_{\text{coarse}}} ||\mathbf{p} + w_{\text{coarse}}\mathbf{b} - \mathbf{p}^{\prime}_{\text{gt}}||^2_2.
\end{aligned}
\end{equation}
Besides, we regularize the predicted baseline vector $\mathbf{b}$ with  $l_{\text{reg}}$ to align it with the point normal $\mathbf{n}$ per valid point: 
\begin{equation}
\begin{aligned}
l_{\text{reg}}=\sum_{\mathbf{b}} \mathop{\min}_{s\in \{-1,1\}}||s\mathbf{n}-\mathbf{b}||_1 + ||1-s\mathbf{n}^T\mathbf{b}||_1. 
\end{aligned}
\end{equation}
The final fitting loss is given as $l=l_{\text{v}}+l_{\text{r}}+l_{\text{w}}+\lambda l_{\text{reg}}$.

A full object-centric representation $\mathcal{I}$ for each instance is given as $\mathcal{I}=(\mathcal{G}, F)$. 

\noindent \subsection{Transferability of NSGF from Semantic Primitives} \label{Sec:trans}
Given  NSGF $F_{\text{src}}$ of one labeled source object $\mathcal{I}_{\text{src}}$ pre-fitted to its grasp labels, we aim to derive a tailored grasp function $F_{\text{tgt}}$ for an unseen target object $\mathcal{I}_{\text{tgt}}$.

\noindent \textbf{NSGF transfer via semantic primitives.}  We probe the source NSGF $F_{\text{src}}$ using semantic primitives $\mathcal{S}_{\text{src}}$ because they greatly help to sample a small yet informative subset of grasps that can express the valid domains of the NSGF. Specifically, each source primitive center $\mathbf{s}_{\text{src}}^j$ is assigned with $N_{j}$ valid grasps decoded from $F_{\text{src}}$ ($N_{j} \leq 5$ for the trade-off between accuracy and speed). The criteria to assign primitive labels for grasps are based on the distance between the left (gripper finger) contact points and the primitive centers. $F_{\text{src}}$ is hence approximated as $\overline{\mathcal{F}}_{\text {src}}$ with: 
\begin{equation}
\begin{aligned}
   \overline{\mathcal{F}}_{\text {src}} &= \{ \overline{\mathcal{F}}_{\text {src}}^j ~|~ j \in \{ 1,..., N_{\mathcal{S}}\} \}, 
\\
   \overline{\mathcal{F}}_{\text {src}}^j &= \{ {\mathbf{g}_{\text{src}}^{j,k}} ~|~ k \in \{1, \ldots, N_{ j}\} \}. 
\end{aligned}
\end{equation}
where $\mathbf{g}_{\text{src}}^{j,k}$ is the $k$-th grasp pose from the $j$-th primitive.

Without loss of generality, we exemplify the approximate NSGF transfer for grasp poses from one primitive, i.e., obtaining $\overline{\mathcal{F}}_{\text {tgt}}^j$ from $\overline{\mathcal{F}}_{\text {src}}^j$. For each source grasp $\mathbf{g}_{\text{src}}^{j,k}$ assigned to $j$-th primitive based on the left contact point, we compute the primitive label for its right contact point, denoted as $j^\prime$. The primitive centers closest to left and right contact points are hence $\mathbf{s}_{\text{src}}^j$ and $\mathbf{s}_{\text{src}}^{j^\prime}$, and the corresponding centers from the target object are $\mathbf{s}_{\text{tgt}}^j$ and $\mathbf{s}_{\text{tgt}}^{j^\prime}$. We first compensate the grasp translation with the averaged primitives offset $\Delta \mathbf{s}$: $\mathbf{g}_{\text{tgt}}^{j,k}[\mathbf{t}] = \mathbf{g}_{\text{src}}^{j,k}[\mathbf{t}] + \Delta \mathbf{s}, \Delta \mathbf{s} = (\mathbf{s}_{\text{tgt}}^j - \mathbf{s}_{\text{src}}^j + \mathbf{s}_{\text{tgt}}^{j^\prime} - \mathbf{s}_{\text{src}}^{j^\prime}) / 2 $. Then we compute the two gripper finger contact points' normals $\mathbf{n}_{\text{tgt}}^{j,k}, \mathbf{n}_{\text{tgt}}^{j,k^\prime}$ on the target object using the grasp $\mathbf{g}_{\text{tgt}}^{j,k}$ translated with  $\Delta \mathbf{s}$, and align the grasp baseline vector computed from the rotation of $\mathbf{g}_{\text{tgt}}^{j,k}$ to $\mathbf{n}_{\text{tgt}}^{j,k}-\mathbf{n}_{\text{tgt}}^{j,k^\prime}$ to respect the antipodal principle.
We repeat this process for other primitives with valid grasps and acquire an approximated NSGF $\overline{\mathcal{F}}_{\text {tgt}}$ (\textit{c.f.} Approx. NSGF in Fig. \ref{Fig: pipeline}-C). As the geometric estimation module provides accurate geometry, we also feed the object mesh $\mathcal{M}_{\text{tgt}}$ and $\overline{\mathcal{F}}_{\text {tgt}}$ to a GPU-accelerated parallel simulator \cite{makoviychuk2021isaac} with free-floating grippers to filter out unstable grasps and acquire a better approximation, $\hat{\mathcal{F}}_{\text {tgt}}$, for the target object's NSGF. Finally, we load the pre-fitted weight of the source NSGF $F_{\text {src}}$ and fit new NSGF for the target object using samples from $\hat{\mathcal{F}}_{\text {tgt}}$ with much fewer iterations (5 times less). 

\noindent \textbf{Inference.} We uniformly sample 5k points on the object surface and feed them in parallel to the transferred NSGF $F_{\text{tgt}}$ to obtain all grasp configurations. We further select the valid grasp poses with the following criteria: (1) predicted as valid ($q > 0$); and (2) kinematically reachable without collision to the scene. We rank them using the shape-aware grasp confidence based on the gradient response introduced in Sec. \ref{Sec:geoinf} (\textit{c.f.} "Shape-aware grasp confidence" in Fig.~\ref{Fig: pipeline}-A), and choose the highest-ranked one.

\subsection{Implementation Details} \label{Sec:implement}
\noindent \textbf{Data preparation.} The 3D models are provided by ShapeNet repository \cite{shapenet2015} with shape augmentation from \cite{wen2022transgrasp}. This resulted in 1,548 mugs, 1,296 bowls, and 1,275 bottles for training. 
We use 3D models with their rendered depth data to train the shape completion network, and 3D models to train the part segmentation network for semantic primitives. 
The source objects with grasp annotations, one per category, are picked randomly from the ACRONYM dataset \cite{acronym2020}. 

\begin{figure*}[t!]
    \centering
    \includegraphics[width=0.9\textwidth]{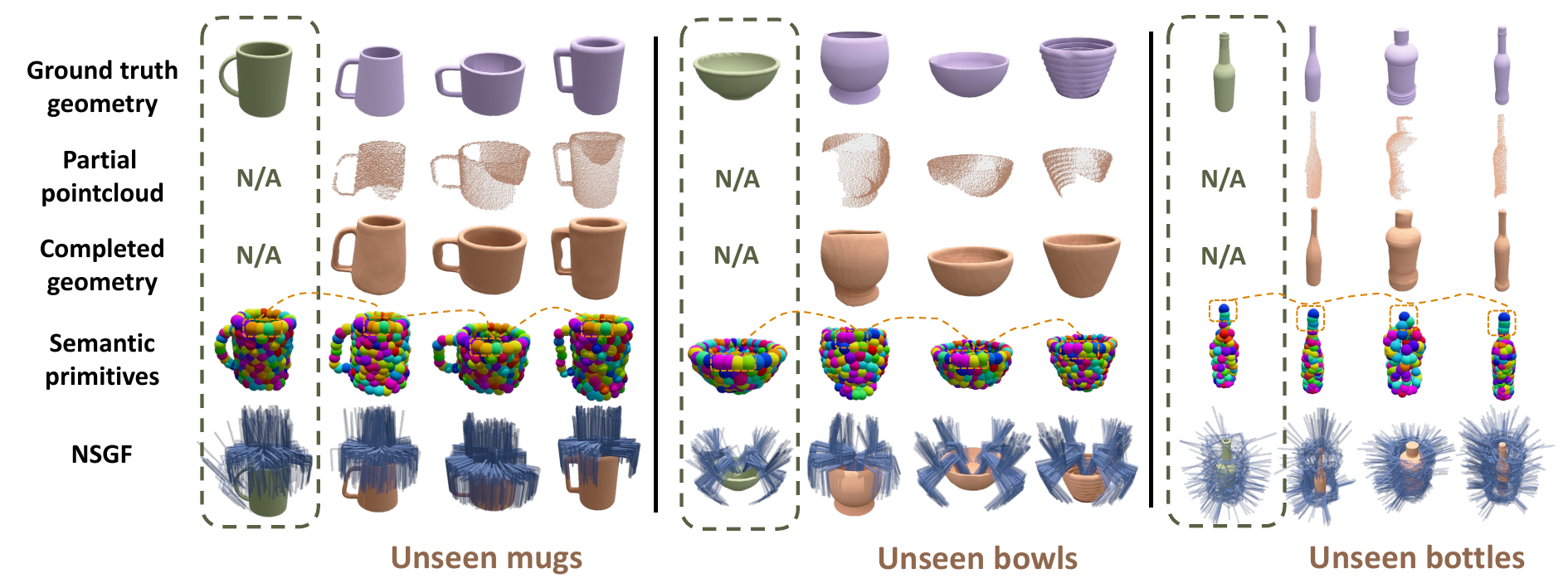}
    \caption{Qualitative results for each category tested in simulations. Source objects with grasp annotations are marked within the green dotted boxes. Their NSGFs are fitted with valid labels. For every unseen object, we visualize its ground-truth geometry (inaccessible for unseen objects),  completed geometry inferred from partial pointcloud, semantic primitives, and the transferred NSGF. We highlight the semantic primitives-based correspondence among objects with orange dotted boxes and lines (zoom in for details). Note here we intentionally down-sample the grasps inferred by NSGF by a factor of 10 compared to the inference time so that the actual ground-truth geometry is visible.}

    \label{Fig: qualitative}
    \vspace{-1.7\baselineskip}
\end{figure*}

\noindent \textbf{Network architecture.} We use an off-the-shelf object detector \cite{rusu2011point}  and pose estimator \cite{li2023generative}. For the shape completion network, we adopt the same architecture as \cite{xu2022learning}.
The semantic primitives' segmentation network is modified from a 3D-GCN \cite{lin2020convolution} with 2,048 points as input, and outputs one-hot logit vectors for 256 primitive labels. The NSGF is designed using SIREN backbone \cite{sitzmann2019siren}, where an 8-layer SIREN network takes a 3-dimensional point coordinate and a 16-dimensional geometric feature trilinearly interpolated from the shape completion network as input and outputs a 128-dimensional feature. Then the feature is fed to two individual 4-layer SIREN networks to output 6-dimensional rotation vectors \cite{zhou2019continuity} and the coarse grasp width. An additional 4-layer SIREN network is used to predict the grasp validity.

\noindent \textbf{Training protocol.} Training of the shape completion network follows \cite{xu2022learning}. The semantic primitives' segmentation network is trained for 10k iterations with a batch size of 12 using Adam \cite{kingma2014adam} with a learning rate of 0.0005. The source object's NSGF fitting takes 200 iterations and uses Adam \cite{kingma2014adam} with a learning rate of 0.0001. $\lambda=0.1$ is set in the fitting loss $l$.

\noindent \textbf{Transfer protocol.} NSGF transfer to the target object loads the pre-fitted weights of the source object's NSGF and is only trained for 40 iterations. Both NSGF fitting and transfer are fed with 2k points per iteration.

\section{Experiments}
\label{sec:experiments}

In this section, we aim to understand the effectiveness of our developed framework. Three different household object categories are tested: Mugs, Bowls, and Bottles. 
We use the same splits and annotated objects as \cite{wen2022transgrasp}. 

\subsection{Experimental Setups}
\begin{figure}[t!]
    \centering
    \includegraphics[width=0.9\columnwidth]{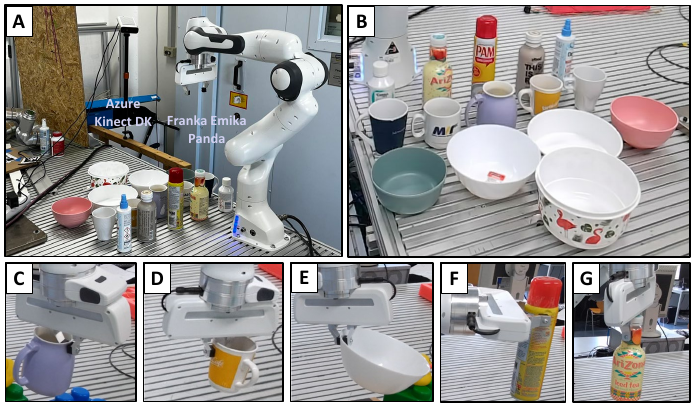}
    % \vspace{-0.8\baselineskip}
    \caption{(A) Physical setup for real robot experiments. (B) Tested objects, five instances per category. (C)-(G) Examples of successful grasps.}

    \label{Fig: real}
    \vspace{-1.7\baselineskip}
\end{figure}

\noindent \textbf{Testing environment.} 
The simulator is built on top of IsaacGym \cite{makoviychuk2021isaac} to evaluate our method and several baselines. The real robotic system consists of a 7-DoF Franka Panda robot and an Azure Kinect DK RGB-D camera (\textit{c.f.} Fig.~\ref{Fig: real}-A). 

\noindent \textbf{Baselines.} For a fair comparison, the baseline methods shall be able to generate sufficiently dense grasp configurations for each object as ours. To this end, our framework is compared against the following representative works: 6-DoF GraspNet (6DGN) \cite{mousavian20196}, Contact GraspNet (CGN) \cite{sundermeyer2021contact} and TransGrasp \cite{wen2022transgrasp}. 6DGN \cite{mousavian20196} and CGN \cite{sundermeyer2021contact} are trained with grasp annotations and agnostic to categories. TransGrasp \cite{wen2022transgrasp} uses one labeled instance and leverages categorical priors as ours. As our generated grasps are defined as a continuous function, we uniformly select 5k points on each object and query their grasp validity and poses from their own NSGFs. We encounter difficulties in comparing our work with other notable works, such as NGDF \cite{weng2023neural}, primarily because it requires dense grasp annotations for all objects during training to establish grasp manifolds. Additionally, it lacks a robust strategy for accurately transferring grasps across instances in a label-efficient manner.

\noindent \textbf{Evaluation protocols.} Our grasping evaluation procedures are categorized into two different setups.

\begin{myitem}
\item {\textbf{Omni-grasp}}: In this setup, we report the success rate and the size of all generated grasps for each object so as to investigate both the reliability and the density of the generated results. We compute the success rate of one object as the ratio of the successful grasps among all valid grasps as \cite{wen2022transgrasp}. For each category, the success rate for this setup is defined as $s_\text{omni}= \nicefrac{1}{N_\text{cat}}  \sum_{i}^{N_\text{cat}}  \nicefrac {N^i_\text{succ}}{N^i_\text{all}}$  
where $N_\text{cat}$ is the number of instances, $N^i_\text{succ}$ and $N^i_\text{all}$ are the successful grasps and all valid grasps for each object, respectively.

\item {\textbf{Best-grasp}}: In this setup, we follow the commonly used evaluation protocol in previous works by selecting the highest-scored grasp configuration among all candidates and grasping each object once. The success rate is hence defined as $s_\text{best}=  \nicefrac {N_\text{cat}^\text{succ}}{N_\text{cat}}$, where $N_\text{cat}^\text{succ}$ is the number of successfully grasped objects using the selected pose.
\end{myitem}

\noindent A grasp is considered successful if the object is lifted for more than 15 seconds. 

\subsection{Results and Discussions in Simulators}
\begin{table*}[t!]
\centering
% \dho{Hanzhi you need to recheck which one is higher...}

% \vspace{-0.2\baselineskip}
\begin{adjustbox}{width=0.9\textwidth,center}
\begin{tabular}{|c|| c   c   c | c || c   c   c| c |}\hline

\hline
 & \multicolumn{4}{c||}{\textbf{Omni-grasp}} & \multicolumn{4}{c|}{\textbf{Best-grasp}}  \\

 \hline
     & Mug & Bowl & Bottle & Avg. & Mug & Bowl & Bottle & Avg. \\
 \hline

6-DoF GraspNet (6DGN) \cite{mousavian20196} & 37.77\% (463.30) & 53.84\% (968.25) & 76.30\% (1147.17) & 55.97\% (859.57) & 42.85\%	& 68.89\% &	78.57\%	& 63.44\%  \\
Contact GraspNet (CGN) \cite{sundermeyer2021contact} & 68.32\% (58.16)	& 76.65\% (177.73) &	72.84\% (122.91) &	72.60\% (119.60) & 69.04\% &	88.89\% &	76.42\% &	78.12\%  \\
TransGrasp  \cite{wen2022transgrasp} & 88.06\% (612.00) &	68.31\% (\textbf{1000.00}) &	86.10\% (1364.00) &	80.82\% (992.00) & 92.07\% &	78.88\%	 & 88.57\% &	86.51\%   \\
\rowcolor[rgb]{ .8,  .8,  .8} \textbf{Ours} & \textbf{94.48\% (801.85)} &	\textbf{92.13\%} (782.55) &	\textbf{86.75\% (2709.75)} &	\textbf{91.12\% (1431.38)}  & \textbf{95.23\%} &		\textbf{95.56\%	 }&	\textbf{91.42\%} &		\textbf{94.07\%}  \\
\hline \hline
Ours w/o width from completion (A1)   & 89.56\% (794.31)	 & 86.62\% (764.33) &	69.53\% (2627.49) &	81.90\% (1395.37) & - & - & - & -  \\
Ours w/o pre-fitting (A2)    & 91.04\% (806.18) & 89.42\% (668.55) &	83.05\% (2643.02) &	87.83\% (1372.58) & - & - & - & -  \\
Ours w/o simulator filtering (A3) & 91.44\% (906.09) &	87.48\% (1013.52) &	85.09\% (2854.03) &	88.00\% (1591.21) & - & - & - & - \\

% \cline{2-3}
% & Reconstruction & Reconstruction \\
% & and completion & only \\
% \hline\hline
% Chamfer Dist. (L1) ($\downarrow$)   & \textbf{0.0157} & 0.0397 \\ 
% \cline{2-3}
% Completeness  ($\downarrow$)        & \textbf{0.0124} & 0.0597 \\   
% \cline{2-3}
% Normal Consis. ($\uparrow$)         & \textbf{0.8969} & 0.6842  \\ 

\hline
\end{tabular}
\end{adjustbox}
\caption{Grasp success rate results of the two evaluation metrics (omni-grasp and best-grasp) tested in simulators. Rows 1-4: Comparison with other methods in simulation. Rows 4-7: Ablation study in simulation. For omni-grasp evaluation results, the numbers of generated grasp poses are given in brackets.}
\label{Tab:results_sim}
\vspace{-1.2\baselineskip}
\end{table*}

Rows 1-4 of Table \ref{Tab:results_sim} show the evaluation results in the simulator for the two setups, i.e., omni-grasp and best-grasp. 

 \noindent \textbf{Omni-grasp evaluation.} We see that methods trained with large-scale datasets (6DGN \cite{mousavian20196} and CGN \cite{sundermeyer2021contact}) perform poorly in omni-grasp evaluation with fewer grasp candidates (859.57 and 119.60) and lower success rate (55.97\% and 72.60\%). As they aim to identify the optimal grasp, they are expected to rigorously filter out the majority of generated results.

 Moreover, they do not retrieve the geometric information of the objects through 3D completion as ours, so occasionally, the gripper could collide with invisible parts of the objects. Such limitation becomes even more evident when CGN \cite{sundermeyer2021contact} deals with bottles as it has an assumption on the thickness of the objects and thus generates fewer poses than the others. TransGrasp \cite{wen2022transgrasp} greatly improves the performance through topology-aware optimizations for grasp poses with a success rate of 80.82\%. However, the size of their grasp proposals for each object is fixed with the number of labeled grasp poses from the source annotated object (992 on average). In contrast, we only use a much smaller subset of the grasps (364.30 on average) for transfer, thanks to the geometric abstraction from primitive-based shape representation. By further leveraging continuous neural representations, the transferred NSGF can smoothly interpolate between discrete samples and produce denser grasp poses than other baseline methods (1431.38 on average). Even with a considerably larger number of grasp poses for evaluation, our success rate still significantly outperforms the strong baseline (TransGrasp \cite{wen2022transgrasp}) by 10.30\%. Besides the effectiveness of our proposed paradigm for grasp transfer with NSGF representation, we also attribute it to the simulator filtering for unstable grasp poses thanks to the accurate geometry from the geometric estimation module. All these results further showcase the impressive ability of our framework to infer reliable yet dense grasp configurations for a wide range of novel objects. Fig. \ref{Fig: qualitative} further exemplifies the inferred geometry, semantic primitives, and NSGFs for several unseen objects. 
 
 \noindent \textbf{Best-grasp evaluation.} Our approach reports the best performance for all categories compared to the others. One major reason is that our generated grasp poses are highly reliable, as shown in the omni-grasp evaluation. Besides, our shape-aware grasp selection strategy based on gradient response helps avoid grasping object regions with noisy observations or unsmooth regions, e.g., the handles of the mugs, which are usually more prone to grasping failures as we noticed in simulators (\textit{c.f.} "Shape-aware grasp confidence" in Fig. \ref{Fig: pipeline}-A).

\subsection{Real Robot Experiments} 
As it would be practically infeasible to perform omni-grasp evaluation in the real world, we fix the object pose in the workspace and test the five furthest valid grasps based on queried points' distance sequentially for approximate verification. For the best-grasp evaluation, we randomly place each object in five arbitrary positions and execute grasping using the inference strategy we suggested.  We evaluate five different instances per category (\textit{c.f.} Fig.~\ref{Fig: real}-B). Both omni-grasp and best-grasp had five trials per object and 25 trials for each category. For the omni-grasp evaluation, we report an overall success rate of 90.66\%, with Mug, Bowl, and Bottle achieving 23/25, 24/25, and 21/25, respectively. The best-grasp evaluation gives a success rate of 93.33\% with each category being 23/25 (Mug), 25/25 (Bowl), and 22/25 (Bottle). These results reflect a similar pattern as the simulations and further suggest though only a limited amount of data is provided (several 3D object meshes and one with grasp labels), our proposed framework is capable of producing dependable grasp poses for real-world objects regardless of appearance, geometry, and positions (\textit{c.f.} Fig.~\ref{Fig: real}-(C-G)). Currently, our approach takes \textit{ca.} 2.71s to fit NSGF for one target object, which includes parallel simulator filtering. Subsequently, the inference time for the fitted NSGF is 0.0043s. These metrics are evaluated on a single NVIDIA GeForce RTX 3080 GPU.

\subsection{Ablation Study}
We conduct ablation experiments to validate crucial components' contributions in rows 4-7 of Table \ref{Tab:results_sim}. For \textit{Ours w/o width from completion} variant (A1), we remove the precise width calculation via the completed shape; for \textit{w/o pre-fitting} variant (A2), we fit the NSGF for unseen objects from scratch instead of loading the pre-fitted NSGF weights of the source annotated object; for \textit{w/o simulator filtering} variant (A3), we skip the simulator filtering before target NSGF fitting. We highlight the effectiveness of shape completion to obtain reliable grasps. As shown in A1, grasping thick objects like bottles easily fails (decrease from 86.75\% to 69.53\%) because the gripper tends to collide due to non-geometry-aware width prediction. Besides, loading the pre-fitted weight of the NSGF from the source object helps faster convergence with 87.83\% compared to 91.12\% of the full model (\textit{c.f.} A2). Simulator filtering (A3) shows its effectiveness with an increase in the success rate of 3.12\% because the inferred shape is sufficiently accurate so it aids in eliminating false positive samples.

\section{Conclusion}
\label{sec:conclusion}

In this work, we propose \coolname{}, a framework to infer dense yet reliable grasp configurations for unseen objects and requires only one annotated object and a single-view pointcloud.
With our \textit{Neural Surface Grasping Fields} formulation, grasp configurations can be effectively embedded on the object surface. This formulation further eases the transfer effort for continuous function-based grasp representation. Using the semantic primitives learned in an unsupervised fashion, we successfully translate smoothly distributed configurations encoded in the grasp function from one single annotated object to several novel instances. The effectiveness is validated through extensive grasping experiments in both simulations and the real world.  

For limitations, the transferred NSGFs could fail to produce grasp configurations for certain regions of some instances, e.g., handles of the mugs. We observe this is occasionally caused by the simulator filtering step. \hl{Furthermore, our framework relies on the fidelity of the physics simulator for seamless sim-to-real transfer. Hence, utilizing a less precise simulator could lead to performance degradation.} In future work, we aim to enhance the speed using techniques like vectorized training \cite{functorch2021}, thereby aligning with low-latency demands.

\clearpage
\addtolength{\textheight}{-12cm}  

%%%%%%%%%%%%%%%%%%%%%%%%%%%%%%%%%%%%%%%%%%%%%%%%%%%%%%%%%%%%%%%%%%%%%%%%%%%%%%%%
% \clearpage
\bibliographystyle{IEEEtran}
\bibliography{literature}
\end{document}